# Cut-and-Paste Dataset Generation for Balancing Domain Gaps in Object Instance Detection

**Woo-han Yun[1, 2], Taewoo Kim[3], Jaeyeon Lee[1], Jaehong Kim[1], and Junmo Kim[4]**
[1]Intelligent Robotics Research Division, ETRI, Republic of Korea
[2]Robotics Program, KAIST, Republic of Korea
[3]Department of Computer Software and Engineering, Korea University of Science and Technology, Republic of Korea
[4]School of Electrical Engineering, KAIST, Republic of Korea

Corresponding author: Junmo Kim (e-mail: junmo.kim@kaist.ac.kr).

This work was supported by the Institute of Information & communications Technology Planning & Evaluation (IITP) grant funded by the Korea government (MSIT) (No. 2017-0-00162, Development of Human-care Robot Technology for Aging Society)

**ABSTRACT** Training an object instance detector where only a few training object images are available is a challenging task. One solution is a cut-and-paste method that generates a training dataset by cutting object areas out of training images and pasting them onto other background images. A detector trained on a dataset generated with a cut-and-paste method suffers from the conventional domain shift problem, which stems from a discrepancy between the source domain (generated training dataset) and the target domain (real test dataset). Though state-of-the-art domain adaptation methods are able to reduce this gap, it is limited because they do not consider the difference of domain gaps of foreground and background. In this study, we present that the conventional domain gap can be divided into two sub-domain gaps for foreground and background. Then, we show that the original cut-and-paste approach suffers from a new domain gap problem, an unbalanced domain gaps, because it has two separate source domains for foreground and background, unlike the conventional domain shift problem. Then, we introduce an advanced cut-and-paste method to balance the unbalanced domain gaps by diversifying the foreground with GAN (generative adversarial network)-generated seed images and simplifying the background using image processing techniques. Experimental results show that our method is effective for balancing domain gaps and improving the accuracy of object instance detection in a cluttered indoor environment using only a few seed images. Furthermore, we show that balancing domain gaps can improve the detection accuracy of state-of-the-art domain adaptation methods.

## I. INTRODUCTION

The development of deep learning has brought tremendous advancement to object detection systems and has led to these systems being deployed in real-world tasks, such as autonomous driving, visual surveillance, medical imaging, and robotics. Consider you build an object detection function for a robot to assist you in cooking or woodworking. This robot has to recognize not only object categories, such as bottles, dishes, cups, and hammers, but it also has to distinguish different object instances within the same category, such as two different bottles. This type of object detection is called object instance detection.

With the help of recent successes of deep convolutional neural networks (CNNs) in object detection tasks [1]–[3], object instance detection systems may be easily built by following ordinary training steps, collecting a dataset, and learning the system with powerful parallel processing hardware. However, in the instance detection problem, it is difficult to collect a large amount of training data for the new object instance, especially in robotic service scenarios. Furthermore, the training dataset should include a range of images captured under various circumstances, such as different viewpoints, illuminations, and backgrounds, with their annotated bounding boxes. Thus, extensive data need to be collected.

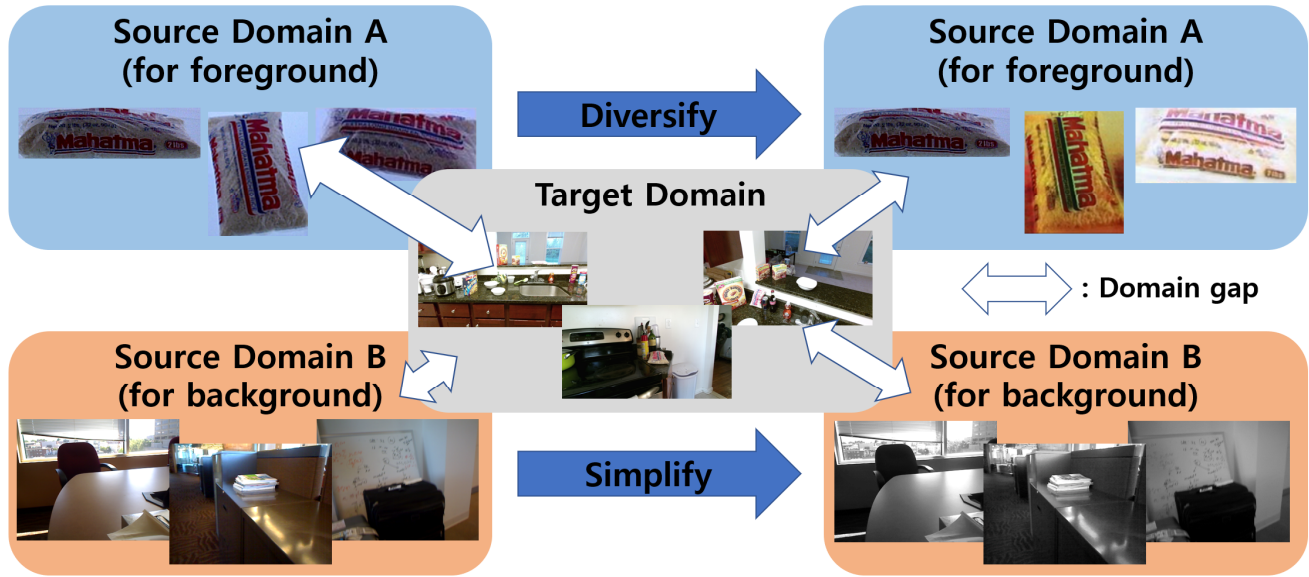

**Figure 1.** As opposed to the ordinary domain shift problem that has one source and one target domain, the cut-and-paste approach has separate source domains for the foreground (source domain A) and the background (source domain B), and two domain gaps that are unbalanced. We propose an advanced cut-and-paste method to balance these domain gaps by diversifying the foreground source domain and simplifying the background source domain.

Recent successful approaches for overcoming this problem render scenes and objects using 3D models [4]–[14]. These 3D models can generate diverse images for the training dataset. However, in a household environment, making accurate 3D models is not a simple task. Furthermore, rendering realistic scenes and objects from these 3D models requires considerable effort and professional skill. Moreover, models trained on such a synthetic dataset suffer from the reality gap problem, which means simulated experience is not carried out into the real world, which several approaches have tried to overcome [5], [6], [8]. Another approach to the data generation problem is cut-and-paste dataset generation, which makes a training dataset directly from real images [15], [16] by cutting an object area and pasting it on randomly selected real background images. This approach requires mask information, which can be generated automatically [15] or manually [16], instead of 3D models.

Our study is based on the second approach, the cut-and-paste dataset generation method. For rapid deployment in a working environment, a few object instance images can be collected in a limited environment by capturing them on a table in a monotone lighting environment. On the other hand, it is also possible to prepare background scenes as varied as possible beforehand using public datasets. These object images and background scenes are not collected in a test environment.

Like other computer vision problems, this cut-and-paste approach also suffers from performance degradation when there is a distribution mismatch between the training domain (or source domain) and the test domain (or target domain).

Domain adaptation methods have been introduced to tackle this domain shift problem by aligning features [39]–[46][61]–[65] and self-training scheme [66]-[70]. However, these approaches require knowledge of the target domain. Furthermore, they assume that the training dataset is collected in one source domain and tested in another target domain. They do not consider the situation in which foreground and background areas are collected in different environments. Because of this reason, the improvement by state-of-the-art domain adaptation methods with the cut-and-paste generation method is limited.

In this study, we first show the cut-and-paste generation method has another type of domain shift problem, because objects and background in the generated training dataset come from different source domains (one for foreground and one for background). We identify that the two domain gaps are unbalanced by comparing two domain gaps from foreground and background source domains to the target domain. Next, we introduce the advanced cut-and-paste dataset generation method to mitigate this gap. Finally, we validate that the proposed methods are helpful in balancing domain gaps and efficient to improve the object detection performance. Balancing domain gaps helps to increase the detection accuracy of state-of-the-art domain adaptation methods. This problem and our solution are shown in Fig. 1. In particular, we notice that the domain gap from each source domain to the target domain is different and unbalanced (in Fig. 1, the size of the arrows represents the size of the domain gap).

Balancing the domain gaps in the cut-and-paste method is important. This leads to performance improvement by reducing the false rejection rate. If the domain gap between the background source domain and the target domain is smaller than the domain gap between the foreground source domain and the target domain (i.e., the training background images are similar to those in the target domain, but foreground objects are not), we can expect that the detector is well-trained on the background but lacks foreground objects. Many variations of foreground objects are considered as a background class, which leads to a high false rejection rate. To the best of our knowledge, this is the first method that balances the domain gap by diversifying the foreground images and simplifying the background images. We can expect that the detector has a higher chance of detecting target objects in the target domain.

Our study makes the following contributions: 1) We identify that unbalanced domain gaps of foreground and background images exists in the original cut-and-paste dataset generation approach. 2) We propose an advanced cut-and-paste dataset generation method that alleviates these unbalanced domain gaps and enhances the detection model to provide a higher detection accuracy with fewer foreground source images.

The remainder of this paper is organized as follows. Section II introduces related works, including studies on object detection, dataset generation, and domain adaptation. We show that the unbalanced domain gap problem exists in the original cut-and-paste dataset generation method and then propose a balanced cut-and-paste dataset generation approach in Sections III and IV, respectively. Section V shows that our method is effective in reducing domain gaps and improving object detection accuracy through various experiments. We conclude our paper in Section VI.

## II. RELATED WORK

Object instance detection has been studied in the field of computer vision and robotics. In early studies, many handcrafted features, such as SIFT (scale-invariant feature transform) [17] and SURF (speeded up robust features) [18] for texture-rich objects, and shape-based methods [19], [20] for texture-poor objects, were used for object instance detection.

Recent detection methods [1]–[3], [13] are based on multilayer CNNs. Because the feature extraction layers of the network have many trainable parameters, these layers are transferred from other networks [21]–[24] trained on massive training datasets, such as ImageNet [25]. Based on these feature extraction layers, specialized architectures for fast object detection have been proposed, such as Faster-RCNN [1], SSD (single shot multibox detector) [2], and Yolo [3].

These object detection algorithms commonly require a large, labeled training dataset. The dataset should have not only numerous images, but also include various variations. This requirement is difficult to satisfy when we apply this system in the home or other similar locations.

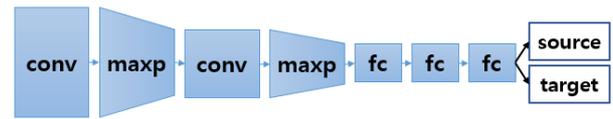

**Figure 2.** Architecture to measure a domain gap between a source domain and a target domain. We use a simple architecture considering the small patch size.

One way to overcome this barrier is to use a rendered image dataset. In this approach, a synthetic image dataset is generated by rendering whole scenes [5], [9]–[11] or composing rendered objects on the real image backgrounds [4], [12]–[14]. These rendered datasets have various object images by rendering 3D models from different viewpoints but struggle with performance degradation from the reality gap between the rendered training dataset and the real test dataset. Recent works [4]–[7] introduced the idea that this domain gap could be reduced by domain randomization.

Alternative approaches [15], [16] use a synthetic dataset generated by placing real segmented object images onto real background images. This approach requires segmentation information such as a mask instead of 3D models. Previous studies obtain this information automatically [15] or manually [16] and place the segmented object on the real backgrounds randomly [15] or according to its scene context [16]. They focused on where and how to place the object instances on the background scenes.

These approaches using synthetic training datasets suffer from domain gaps or domain shift problems because their training domains do not completely coincide with the test domain. To address this problem, domain adaptation is widely studied in computer vision and machine learning [26]–[31]. Recent works introduce domain adaptation methods on deep learning frameworks. Many previous works focused on the object classification problem [32]–[38], [61], [62] and, some studies used generative adversarial networks (GANs) for domain adaptation at the pixel level [43]–[46]. Although less attention has been paid to domain adaptation for object detection tasks, some studies in this area [39]–[42], [63]–[65] have been done. They presented an adaptive structural SVM (support vector machine) for a DPM (deformable part models)-based object detector [39], an adaptive decorrelation approach based on data statistics [40], and subspace alignment on feature representations [41].

More recently, adversarial training is actively used to reduce the domain gap in the deep learning framework. Those works are applying adversarial training at image level and instance level [42], aligning features strongly at a low-level feature and weakly at a high-level feature [63], adding cycle-consistent constraint for preserving identity [64], and progressively reducing domain gap via intermediate domain at image level [65] or score level [62].

Another approach is to use a self-training scheme. Self-training approaches are mainly composed of two-step, getting pseudo labels on the target domain dataset and finetuning the detection model with these pseudo labels. Many works have focused on the first step, how to get believable pseudo labels. They presented weakly-supervised learning by image-level domain transfer and annotation information [66], exploiting temporal cues using a tracker on unlabeled videos is presented [67], relying on pseudo labels of the intermediate domain and the ground-truth of source domain in an imbalanced sampling way [68], using k-reciprocal encoding and clustering algorithm in feature space [69], and treating easy and hard pseudo labels differently [70].

These works commonly require information about the target domain through unlabeled or a low number of target-domain samples. However, the proposed method does not require knowledge of the target domain. Furthermore, domain adaptation methods do not consider the situation where foreground and background areas are collected in different environments. With the help of the proposed method, domain adaptation methods can reach higher detection accuracy by considering the unbalanced domain gap problem if we can access the unlabeled target dataset.

## III. UNBALANCED DOMAIN GAP

Our work is based on cut-and-paste dataset generation. This approach has two source domains, one for foreground objects and one for background areas. This entails two different domain gaps, which could be considered an unbalanced domain gap problem. To compare the domain gaps of the foreground (object instances) and background (background scenes), we use $H$-divergence [42][47], which is designed to measure the divergence between two sets of samples from different domains or distributions. Let $S$ and $T$ be a set of source and target domains, respectively. $x$ is a feature vector of a sample from $S$ or $T$. $h$ is one of the domain classifiers in $H$ to classify $S$ and $T$ given $x$. $H$ is a set of possible domain classifiers. $H$-divergence is defined as follows:

$$d_H(S,T) = 2 \times (1 - \min_{h \in H}(err_S(h(x)) + err_T(h(x)))), \quad (1)$$

where $err_S$ and $err_T$ is the prediction error of domain classifier $h$ given a sample $x$ from the source and target domain, respectively. $H$-divergence implies that the domain distance $d_H(S, T)$ depends on how well the samples of each domain are separable. If the samples are well separable, the prediction error will be low. Therefore, the $H$-divergence is inversely proportional to the prediction errors.

To measure the $H$-divergence in our problem, we prepared the dataset and classifier set $H$ as follows. We first created a dataset containing 10,000 patches with $32 \times 32$ pixels of foreground and background. Foreground patches are randomly cropped in a bounding box of object instances. Background patches are selected from the background area to be IoU (intersection over union) < 0.1 with the bounding boxes of object instances. The considered classifier is a simple CNN with an architecture of conv(k5-f6)-maxp(k2)-conv(k5-f16)-maxp(k2)-fc1(f120)-fc2(f84)-fc3(f2), where

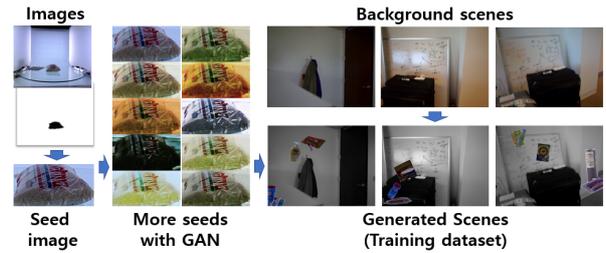

**Figure 3.** Our approach to generating a training dataset. Starting from a set of object images and masks, diverse seed images are generated using GAN and pasted onto the simplified background scene. With these two processes, the domain gaps from two source domains to a target domain are balanced.

k5-f6 is 6 filters with a kernel of $5 \times 5$, as shown in Fig. 2. The ReLU activation function is applied after every conv and fc layer except for the last layer. We used a simple architecture because the patch size was relatively too small for using other heavy architectures, such as VGG [21] or ResNet [23]. We split the dataset into training, validation, and testing sets in a ratio of 70%, 10%, and 20%, respectively. The sets do not share the same background or scene. We selected the best classifier with the best validation accuracy. The $H$-divergence we measured is shown in Table I.

TABLE I
$H$-DIVERGENCE OF FOREGROUND AND BACKGROUND PATCHES BETWEEN SOURCE DOMAIN AND TARGET DOMAIN

|  | Foreground | Background |
| --- | --- | --- |
| $H$-divergence | 1.616 | 1.016 |

From Table I, we see that the domain gaps of the foreground and background are unbalanced, 1.616 and 1.016, respectively. This means that foreground patches are easily distinguishable between the training set (source domain) and the test set (target domain), but the background is relatively hard to distinguish. If we have information about the target domain in advance, we could use this information to reduce the domain gap by reducing $H$-divergence [42]. In this work, we propose to balance this unbalanced domain gap without any information about the target domain.

## IV. BALANCING DOMAIN GAP

An overview of our balanced cut-and-paste dataset generation approach is shown in Fig. 3. The overall process mainly follows previous research on cut-and-paste dataset generation [15], but some steps are modified to balance the domain gaps of foreground and background images. We briefly introduce the steps 1–6 [15], and then focus on steps 4 and 5, which are the ones we mainly modified. Our method consists of collecting images of object instances and background scenes, processing the collected images, and pasting object areas onto background scenes.

1) COLLECT OBJECT IMAGES

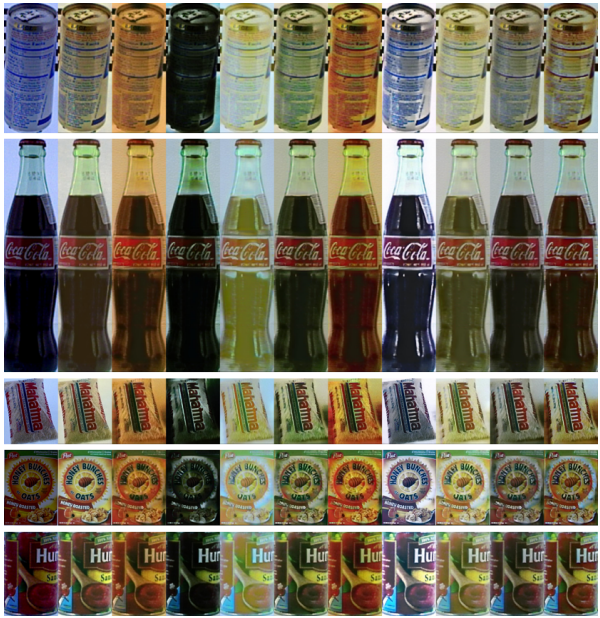

**Figure 4.** Seed images generated using MUNIT [46]. The first column is input images from the BigBIRD Dataset. The second column contains images generated without style codes. The other columns show randomly sampled styles. Images in the same column were generated from same style code.

Photos of object instances are taken from each surface and corner. These images are used as seed images to create a training dataset.

2) COLLECT BACKGROUND SCENE IMAGES
These images are used as a background. This step can be replaced by preparing a public dataset.

3) PREDICTION OF FOREGROUND MASK
This step makes masks of object instance images. These masks are used to segment the object instances from the images.

4) SIMPLIFYING THE BACKGROUND IMAGES
To expand the domain gap of the background images, the image quality of these images is deteriorated using an image processing technique.

5) DIVERSIFYING THE FOREGROUND IMAGES
To reduce the domain gap of foreground images, more diverse foreground images that can be used as seed images are generated.

6) PASTE OBJECT INSTANCES INTO BACKGROUND SCENES
Finally, segmented object instances are pasted onto randomly chosen background images. To reduce local artifacts at the object boundaries, the method in [15] are adopted. Data augmentation is also applied to simulate variations, such as translation, rotation, and scaling of object instances.

*A. Simplifying background images*
Under the assumption that the background source dataset is prepared in advance and collected from diverse environments, we decided to shrink its domain by reducing its diversity. To do this, we deteriorate background images with image processing methods, such as Gaussian blurring (to attenuate its high-frequency information), graying, and quantizing color to 8 bits (to lose its color information). From this simple image transformation, we expect that this step causes background images to be unrealistic and the domain gap of the background to be enlarged (by shrinking its background source domain). Simply, color-to-gray transformation turns various possible color images that have a certain appearance into one gray image. This reduces its diversity and increases its domain gaps. In our experiment, we validated that this process expands the domain gap.

*B. Diversifying foreground images*
Contrary to the background dataset, foreground objects are collected in a limited situation. Therefore, we expand the foreground source domain by making the seed images more diverse. We generate diverse seed images using the GAN. As a GAN model, we use a multimodal unsupervised image-to-image translation (MUNIT) model [46]. In this model, the image is decomposed into a content code that is domain-invariant and a style code that is domain-specific. The image could have other styles by combining the content code with other random style codes. This model could be trained without pair-wise datasets in the source and target domains. For training, we used the images in the BigBIRD Dataset and Active Vision Dataset (AVDataset) [53] as the source and target domains, respectively. The object instances we used are not in the GMU Kitchen dataset [54], but in AVDataset, because the GMU Kitchen dataset is used for evaluation. The resulting images are shown in Fig. 4. The results in each column were generated with the same style code. Images in the first and second columns are input images and result images generated without the style code (by giving zero code), respectively. We identify that the trained styles are mainly lighting effects, such as spotlights and the color tones of lights. With the MUNIT model, we can insert various style effects to the object instance image. From this step, we expect that the foreground source domain expands its boundary and the domain gap between the foreground source domain and the target domain is reduced.

## V. EXPERIMENTS
In this section, we show the accuracy improvement of the cut-and-paste dataset-generation method by adopting the steps we discussed in Section IV. A and IV. B. Then, an accuracy comparison with other cut-and-paste dataset generation approaches is presented. We also demonstrate that the unbalanced domain gap is diminished by comparing the original cut-and-paste method and the proposed method. Finally, we apply our method to another target dataset. We first describe the common experimental setups.

1) SOURCE AND TARGET DATASET
We use object instances in the BigBIRD dataset [48] as the foreground source dataset. In the BigBIRD dataset, 125

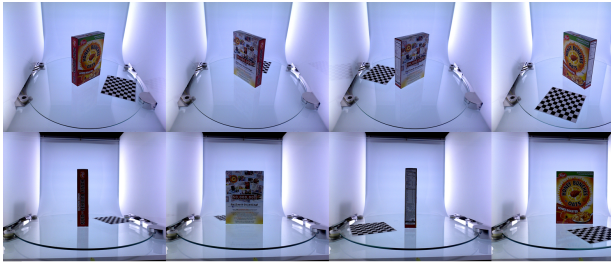

**Figure 5.** Eight seed images of the "honey bunches" class. We select images that show as many faces as possible in one image.

object instances have 600 images taken from different viewpoints, at five elevations and from 360 angles. To reduce the human effort to capture object images, we used 8 seed images for each object. We selected images to show as many faces as possible in one image. The selected seed images are shown in Fig. 5. For the background source dataset, 1548 background images from the UW Scenes dataset [49] were used. We used the GMU Kitchen dataset [54] as the target domain for reporting object instance detection performance and comparison in all experiments, unless otherwise noted. Eleven object instances overlapped on both the BigBIRD dataset (foreground source domain) and the GMU Kitchen Dataset (target domain) [54]. For training the MUNIT model in Section IV. B, we used 26 object instances from AVDataset [53]. We only used object instances that are not included in the GMU Kitchen dataset for training the MUNIT model.

### 2) DATASET GENERATION

To extract the object instance area in a seed image, we used GrabCut [50]. All masks were refined with a hole-filling algorithm to remove any interior holes. To create the training dataset, we used the cut-and-paste method posted at https://github.com/debidatta/syndata-generation. The final generated training dataset has approximately 6000 images with a resolution of 640 × 480 pixels. We found that placing objects in random positions in [15] rarely causes occlusion. To make more occlusions, we selected the positions of the object instances to be near the previous positions of the other objects with a 50% chance. We refer to this version as occV2. We followed the default options in [15] for all other options.

### 3) MODEL AND LEARNING

We used a PyTorch implementation of Faster R-CNN [1] based on VGG-16 [21] at https://github.com/jwyang/faster-rcnn. The network was initialized with weights pretrained on the MSCOCO [55] dataset, then finetuned on each generated dataset. The network was trained for 4 epochs using the SGD+momentum optimizer with a learning rate of 0.001, and a momentum of 0.9. The learning rate was reduced by a factor of 10 after 2 epochs. For consistent evaluation, we fixed all hyperparameters and random seeds across experiments. We resized the images to have a short length of 800 pixels. For training the MUNIT model, we used a PyTorch code at https://github.com/NVlabs/MUNIT with the synthia2cityscape settings. We added Laplacian loss [56], [57] to preserve the detail-enhanced images. The images, resized to 384 × 384 pixels, were used for both the source and target domains. The model was trained for 300,000 iterations with a batch size of 1.

### 4) EVALUATION

Accuracy was measured using mean average precision (mAP) at an IoU of 0.5. We report the mAP calculated using the PASCAL VOC MATLAB code [58]. Boxes of at least 50 × 30 pixels are used as an evaluation for consistency with previous works [15], [16].

### A. Simplifying background images

To see the effect of simplification on background images, we applied three image processing techniques, Gaussian blurring, gray conversion, and 8-bit color quantization, to background images. To inspect the effect according to the complexity of the background images, we tested more images with another public dataset, COCO [55], which includes more diverse images than the UW Scenes dataset by including indoor/outdoor and various object images. We selected 1,500 images from the COCO dataset and used them randomly as background images.

TABLE II
EVALUATION OF SIMPLIFYING SOURCE BACKGROUND IMAGES

| Simplification method | mAP | |
|---|---|---|
| | UW Scenes | COCO |
| w/o processing | 67.2 | 42.4 |
| Gaussian blur | 72.7 | 74.3 |
| Gray | 75.3 | 74.7 |
| 8-bit quantization | 72.8 | 65.3 |

Table II shows the evaluation results. We observe that the detection accuracy depends on the complexity of the source background dataset. Without any processing on the background images, the result using UW Scenes background images is higher than that using COCO background images. We expect that the COCO background includes too many types of scenes, such as foods, vehicles, animals, and both indoor and outdoor scenes. This diversity of background images hinders the realization of a proper gap between the source background domain and the target domain compared to the gap between the source foreground and the target domain. This leads to false rejection of many foreground objects. Regardless of the background complexity, all simplification methods contribute to improving the accuracy of object detection. In particular, removing color information by converting background images to grayscale increased the method's accuracy by the largest margin on each background dataset.

### B. Diversifying foreground images

Based on the result (Gray, in Table II) in Section V. A, we evaluated the methods for diversifying foreground images in terms of object detection performance. These generate more occlusion (marked as occV2) and use more seed images generated by GAN (marked as GAN). In Table III, we see that generating more occlusion situations (overlapping with previous object positions with 50% probability) leads to higher accuracy by a small margin. However, this does not mean that higher occlusion always leads to higher accuracy, because we observe that the performance decreases when we generate more occlusions. When we train the model on only GAN-generated seed images, the performance degraded by a small margin (Gray-occV2-GAN in Table III). With the training images generated from both original seed images and GAN-generated seed images at an equal ratio, we obtained the highest accuracy (Gray-occV2-HalfGAN in Table III).

TABLE III
EVALUATION OF DIVERSIFYING SOURCE FOREGROUND IMAGES

| Diversifying method | mAP |
|---|---|
| Gray | 75.3 |
| Gray-occV2 | 76.4 |
| Gray-occV2-GAN | 70.5 |
| Gray-occV2-HalfGAN | 78.1 |

*C. Comparison with other methods*

In Table IV, we compare our results with other cut-and-paste results in [15], [16] and the domain adaptation method in [63]. We note that [16] used 360 seed images and approximately 7000 background images from the NYUD v2 dataset [59]. They used global structure information (e.g., objects on flat surfaces) to paste images. Study [15] randomly pasted object images from 600 seed images onto the background images from the UW Scenes dataset [49]. To obtain the result of the domain adaptation without considering unbalanced domain gaps, we used the implementation in [63] on the generated dataset set in [15]. We also report the results of the trained model with labeled real images in the target domain (the GMU Kitchen dataset).

TABLE IV
COMPARISON WITH OTHER CUT-AND-PASTE METHODS AND TRAINING DATASETS

| Dataset Generation method | Used dataset for training and dataset generation | | | # of seed images | mAP |
|---|---|---|---|---|---|
| | fg src* | bg src* | tgt* | | |
| - | × | × | ○ | - | 86.2 |
| - | ○ | × | × | 8 | 49.6 |
| [15] | ○ | ○ | × | 600 | 76.2 |
| [16] | ○ | ○ | × | 360 | 51.7 |
| [15] | ○ | ○ | × | 8 | 67.2 |
| [15]+[63] | ○ | ○ | △** | 8 | 72.5 |
| [15]+[63]+[68] | ○ | ○ | △** | 8 | 79.0 |
| Ours | ○ | ○ | × | 8 | 78.1 |
| Ours+[63] | ○ | ○ | △** | 8 | 79.4 |
| Ours+[63]+[68] | ○ | ○ | △** | 8 | 81.6 |

* tgt: target, src: source, fg: foreground, bg: background.

** △: unlabeled target dataset is used for domain adaptation

The best accuracy comes from the model trained on the labeled target images. If we train the detection model with only a foreground source dataset (without dataset generation), we obtain a poor result of 49.6%. With the original cut-and-paste method using 8 seed images, we get 67.2% detection accuracy. This accuracy is lower than that in [15] and higher than that in [16], even though they used as many seed images to generate a training dataset, 600 viewpoint images in [15], and 360 viewpoint images in [16], for each object instance. We assume that this performance improvement with fewer seed images stems from the differences in the detailed options of the training models and the background images and masks.

Compared with the original cut-and-paste method using 8 seed images, our method increased the detection accuracy by 10.9% (67.2% → 78.1%). This is the best accuracy among the dataset generation methods using foreground and background datasets without knowledge of the target domain. If knowledge of the unlabeled target dataset is available, unsupervised domain adaptation methods could be applicable. We applied two unsupervised domain adaptation methods using feature alignment [63] and self-training [68] on both datasets. Both adaptation methods helped to improve detection accuracy. In detail, feature alignment [63] improved the detection accuracy to 72.5% with the original cut-and-paste method and to 79.4% with our proposed method. The self-training method [68] additionally increased to 79% and 81.6% with each dataset generate method, respectively. Even though applying feature alignment on the original cut-and-paste dataset increased its detection accuracy to 72.5%, that was less than the result of our method without feature alignment, 78.1%. This shows that balancing domain gaps is an important problem than reducing the domain gap without considering the balance of domain gaps. Self-training method improved by a big margin in both datasets, 6.5% and 2.2%, respectively. This is because the self-training method relies on the target dataset than the source dataset. But, more accurate pseudo labels generated from our method lead to higher detection accuracy than the original approach. From these experiments, we found that balancing domain gaps is important as it is and boosts the performance of existing domain adaptation methods.

In Fig. 6, the detection results of the original cut-and-paste methods (left) and ours (right) are illustrated. The detection model trained on our generated dataset is more robust to variations in object instances, such as occlusion and illuminations (red boxes in Fig. 6). It also presents tighter bounding boxes for objects (yellow boxes). Both methods have false acceptance cases (purple boxes), but our method is more vulnerable to false acceptance errors.

*D. Domain gaps*

To identify the change in domain gaps before and after our methods, we measured the domain gap using *H*-divergence in Section III and plotted the data points with t-SNE [60]. To plot the distribution of data, we used feature vectors at the fc2 layers. Table V shows that the domain gap of the foreground was reduced from 1.6 to 1.3, and the gap of the background was increased from 1.0 to 1.5. As a result, the gap in the domain gaps of the foreground and background was reduced from 0.6 to 0.12. In Fig. 7, we identify the shift in the domain gap. The foreground data points before and after using our method are in Fig. 7 (a) and (c), and the background data points are in Fig. 7 (b) and (d). From Fig. 7 (a) to (c), we can see that the data points become more overlapped. Conversely, from Fig. 7(b) to (d), we see that the background data becomes more separated.

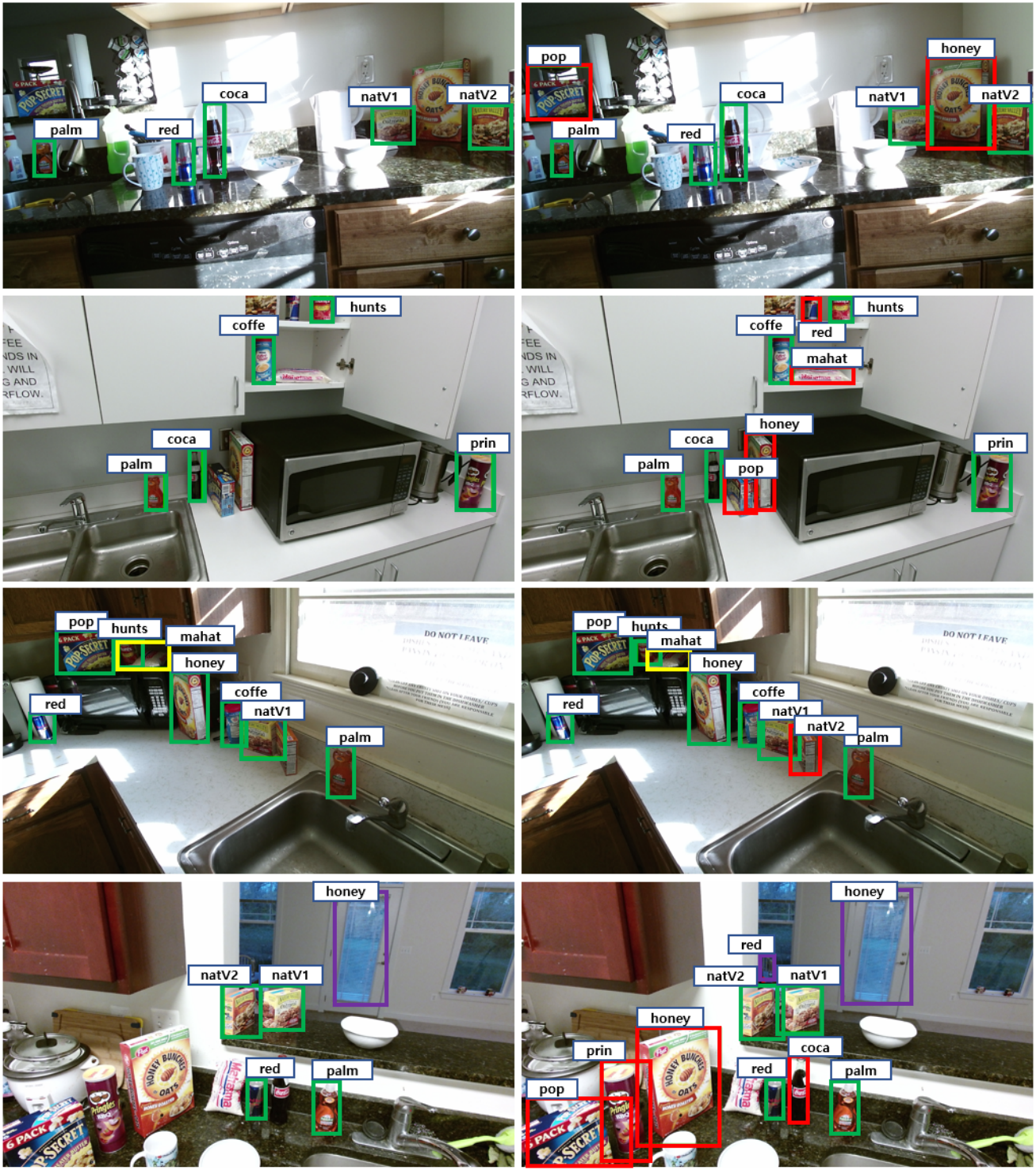

**Figure 6.** Detection results of the original cut-and-paste method (left) and our final method (right). The detection model trained on our generated dataset is more robust to variations in the object instances, such as occlusions and illuminations, with tight bounding-boxes (See red and yellow boxes). However, our method is more vulnerable to false acceptance error (purple boxes).

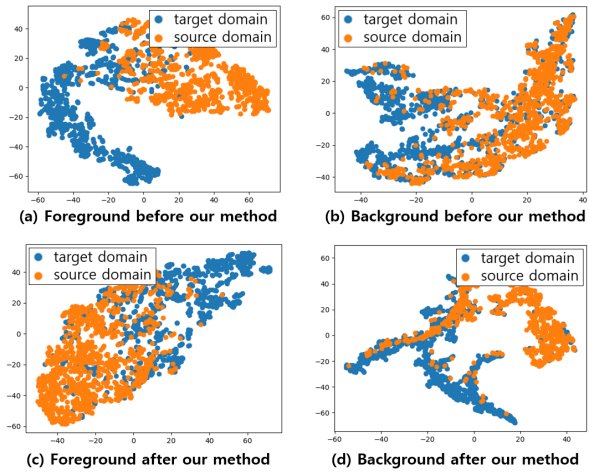

**Figure 7.** t-SNE visualization of feature vectors at the fc2 layers. (a) and (b) are the original data of foreground and background; (c) and (d) are the data after applying our algorithm.

We expect that the process for foreground images generates diverse foreground images in the generated dataset. This expands its boundary of foregrounds of the source domain and reduces its gap with the target domain. Conversely, the process for background images, such as graying and blurring, shrinks the data distribution (for example, if we blur images with an infinite kernel size, we get some similar images with single values). This pulls its background boundary in the source domain and expands its gap with the target domain.

TABLE V
DOMAIN GAP USING $H$-DIVERGENCE OF FOREGROUND AND BACKGROUND PATCHES WITH AND WITHOUT OUR METHOD

| $H$-divergence | Foreground | Background | Gap |
|---|---|---|---|
| Before | 1.616 | 1.016 | 0.60 |
| After | 1.336 | 1.456 | 0.12 |

*E. Evaluating another target domain*

To evaluate the generalization capability of our methods, we evaluated our approach on another target dataset. We trained the detection system on our dataset (Gray-occV2-HalfGAN in Table III) and all images of the GMU Kitchen dataset, and evaluated it on AVDataset [53]. The AVDataset includes color and depth images of 33 object instances. The dataset was first released as 17,556 images in 9 scenes, and we evaluated this release in our test. We report the results of six object instances overlapped between the GMU Kitchen and AVDataset. We also present the accuracy by varying the number of real images from the GMU Kitchen dataset. For this test, we did not train further on AVDataset.

In Table VI, the result with models trained on the dataset generated with only eight seed images is higher than the result with 1% of the real GMU Kitchen dataset (35.8% vs. 24.5%). The more real images we add in the training dataset, the higher the accuracy we can obtain (35.8% → 49.2% → 53.3% → 55.4%). Interestingly, the result with our generated dataset and 10% real images shows higher accuracy than the model trained on 100% real images (53.3% vs. 52.5%). This means that we can obtain a similar detection performance with only a small portion of real images and fewer seed images if we apply our methods.

TABLE VI
EVALUATION OF THE AVDATASET IN CROSS-DOMAIN SETTINGS

| Ratio of Real | Trained only real GMU dataset | Trained both real GMU and our dataset |
|---|---|---|
| 0% | - | 35.8 |
| 1% | 24.5 | 49.2 |
| 10% | 48.2 | 53.3 |
| 100% | 52.5 | 55.4 |

## VI. CONCLUSION

In this work, we presented methods to generate training images from seed images. We identified that the domain gap is unbalanced when we paste the seed images onto background scenes to generate a training dataset. To balance the domain gap, we introduced methods for diversifying the foreground images and simplifying the background images. In the experiments, we showed that our method is effective for balancing domain gaps and is helpful in improving the accuracy of object instance detection in cluttered indoor environments with limited seed images.

The proposed method assumed that a few object images collected in a limited environment and diverse background images are prepared for training dataset generation. This assumption is valid if we consider building an object detector in a limited environment and time, e.g. building an object instance detector function for a robot to assist you in cooking or woodworking with a few images. In this case, the object instance detector trained with the original cut-and-paste method has suffered from false rejection error because the diverse background and limited foreground make the domain gaps are unbalanced and the trained model can misclassify the object instance area as a background. If we build the object instance detector with the help of our method, the domain gaps of foreground and background are balanced and the detector can reduce the false rejection error.

However, the opposite case is also possible where we have much more diverse object instance images and a few background images. In this case, the proposed method will not work as intended. To overcome this limitation, our approach has to dynamically gauge the domain gaps of foreground and background between the source domain and the target domain and diminish both gaps in a balanced way. In future work, we plan to build an end-to-end deep learning framework that generates a dataset by considering the $H$-divergence gap as a metric.

## REFERENCES

[1] S. Ren, K. He, R. Girshick, and J. Sun, "Faster R-CNN: Towards Real-Time Object Detection with Region Proposal Networks," *IEEE Trans. Pattern Anal. Mach. Intell.*, vol. 39, no. 6, pp. 1137–1149, 2017.

[2] W. Liu *et al.*, "SSD: Single Shot MultiBox Detector," in *Proc. ECCV*, 2016, pp. 21–37.

[3] J. Redmon and A. Farhadi, "YOLO9000: Better, Faster, Stronger," in *Proc. CVPR*, 2017, pp. 6517-6525.

[4] Y. Movshovitz-Attias, T. Kanade, and Y. Sheikh, "How useful is photo-realistic rendering for visual learning?," in *Proc. ECCV*, 2016, pp. 202–217.

[5] J. Tobin, R. Fong, A. Ray, J. Schneider, W. Zaremba, and P. Abbeel, "Domain Randomization for Transferring Deep Neural Networks from Simulation to the Real World," in *Proc. IROS*, 2017, pp. 23–30.

[6] J. Tremblay, T. To, B. Sundaralingam, Y. Xiang, D. Fox, and S. Birchfield, "Deep Object Pose Estimation for Semantic Robotic Grasping of Household Objects," in *Proc. CoRL*, 2018.

[7] J. Tremblay *et al.*, "Training deep networks with synthetic data: Bridging the reality gap by domain randomization," in *Proc. CVPR Workshop*, 2018, pp. 1082–1090.

[8] A. Prakash *et al.*, "Structured Domain Randomization: Bridging the Reality Gap by Context-Aware Synthetic Data," in *Proc. ICRA*, 2019.

[9] A. Rozantsev, V. Lepetit, and P. Fua, "On Rendering Synthetic Images for Training an Object Detector," *Comput. Vis. Image Underst.*, vol. 137, pp. 24–37, Aug. 2015.

[10] S. R. Richter, V. Vineet, S. Roth, and V. Koltun, "Playing for Data: Ground Truth from Computer Games," in *Proc. ECCV*, 2016.

[11] A. Handa, V. Patraucean, V. Badrinarayanan, S. Stent, and R. Cipolla, "SceneNet: Understanding Real World Indoor Scenes With Synthetic Data," 2015. [Online]. Available: arXiv:1511.07041v2.

[12] A. Gupta, A. Vedaldi, and A. Zisserman, "Synthetic Data for Text Localisation in Natural Images," in *Proc. CVPR*, 2016.

[13] X. Peng, B. Sun, K. Ali, and K. Saenko, "Learning Deep Object Detectors from 3D Models," in *Proc. ICCV*, 2015.

[14] H. Su, C. R. Qi, Y. Li, and L. Guibas, "Render for CNN: Viewpoint Estimation in Images Using CNNs Trained with Rendered 3D Model Views," in *Proc. ICCV*, 2015.

[15] D. Dwibedi, I. Misra, and M. Hebert, "Cut, Paste and Learn: Surprisingly Easy Synthesis for Instance Detection," in *Proc. ICCV*, 2017, pp. 1310–1319.

[16] G. Georgakis, A. Mousavian, A. Berg, and J. Kosecka, "Synthesizing Training Data for Object Detection in Indoor Scenes," 2017. [Online]. Available: arXiv:1702.07836v2.

[17] D. G. Lowe, "Distinctive Image Features from Scale-Invariant Keypoints," *Int. J. Comput. Vis.*, pp. 1–28, 2004.

[18] H. Bay, A. Ess, T. Tuytelaars, and L. Van Gool, "Speeded-Up Robust Features (SURF)," *Comput. Vis. Image Underst.*, vol. 110, no. 3, pp. 346–359, 2008.

[19] S. Hinterstoisser *et al.*, "Multimodal Templates for Real-Time Detection of Texture-less Objects in Heavily Cluttered Scenes," in *Proc. ICCV*, 2011, pp. 858–865.

[20] V. Ferrari, T. Tuytelaars, and L. Van Gool, "Object detection by contour segment networks," in *Proc. ECCV*, 2006, pp. 14–28.

[21] K. Simonyan and A. Zisserman, "Very Deep Convolutional Networks for Large-Scale Image Recognition," in *Proc. ICLR*, 2014, pp. 1–14.

[22] C. Szegedy *et al.*, "Going Deeper with Convolutions," in *Proc. CVPR*, 2015, pp. 1–9.

[23] K. He, X. Zhang, S. Ren, and J. Sun, "Deep Residual Learning for Image Recognition," in *Proc. CVPR*, 2016, pp. 770–778.

[24] A. Krizhevsky, I. Sutskever, and G. E. Hinton, "ImageNet Classification with Deep Convolutional Neural Networks," in *Proc. NeurIPS*, 2012, pp. 1097–1105.

[25] J. Deng *et al.*, "ImageNet: A Large-Scale Hierarchical Image Database," in *Proc. CVPR*, 2009.

[26] B. Kulis, K. Saenko, and T. Darrell, "What You Saw is Not What You Get: Domain Adaptation Using Asymmetric Kernel Transforms," in *Proc. CVPR*, 2011, pp. 1785–1792.

[27] B. Gong, Y. Shi, F. Sha, and K. Grauman, "Geodesic Flow Kernel for Unsupervised Domain Adaptation," in *Proc. CVPR*, 2012, pp. 2066–2073.

[28] B. Fernando, A. Habrard, M. Sebban, and T. Tuytelaars, "Unsupervised Visual Domain Adaptation Using Subspace Alignment," in *Proc. ICCV*, 2013.

[29] B. Sun, J. Feng, and K. Saenko, "Return of Frustratingly Easy Domain Adaptation," in *Proc. AAAI*, 2016, pp. 2058–2065.

[30] R. Gopalan, R. Li, and R. Chellappa, "Domain Adaptation for Object Recognition: An Unsupervised Approach," in *Proc. ICCV*, 2011, pp. 999–1006.

[31] L. Duan, I. W. Tsang, and D. Xu, "Domain Transfer Multiple Kernel Learning," *IEEE Trans. Pattern Anal. Mach. Intell.*, vol. 34, no. 3, pp. 465–479, 2012.

[32] F. M. Cariucci, L. Porzi, B. Caputo, E. Ricci, and S. R. Bulo, "AutoDIAL: Automatic Domain Alignment Layers," in *Proc. ICCV*, 2017, pp. 5077–5085.

[33] M. Ghifary, W. B. Kleijn, M. Zhang, D. Balduzzi, and W. Li, "Deep Reconstruction-Classification Networks for Unsupervised Domain Adaptation," in *Proc. ECCV*, 2016.

[34] Y. Wang, W. Li, D. Dai, and L. Van Gool, "Deep Domain Adaptation by Geodesic Distance Minimization," in *Proc. ICCV Workshop*, 2018, pp. 2651–2657.

[35] Y. Ganin and V. Lempitsky, "Unsupervised Domain Adaptation by Backpropagation," in *Proc. ICML*, 2015.

[36] O. Sener, H. O. Song, A. Saxena, and S. Savarese, "Learning Transferrable Representation for Unsupervised Domain Adaptation," in *Proc. NeurIPS*, 2016.

[37] P. Haeusser, T. Frerix, A. Mordvintsev, and D. Cremers, "Associative Domain Adaptation," in *Proc. ICCV*, 2017, pp. 2784–2792.

[38] S. Motiian, M. Piccirilli, D. A. Adjeroh, and G. Doretto, "Unified Deep Supervised Domain Adaptation and Generalization," in *Proc. ICCV*, 2017, pp. 5716–5726.

[39] J. Xu, S. Ramos, D. Vazquez, and A. M. Lopez, "Domain Adaptation of Deformable Part-Based Models," *IEEE Trans. Pattern Anal. Mach. Intell.*, vol. 36, no. 12, pp. 2367–2380, 2014.

[40] B. Sun and K. Saenko, "From virtual to reality: Fast adaptation of virtual object detectors to real domains," in *Proc. BMVC*, 2014.

[41] A. Raj, V. P. Namboodiri, and T. Tuytelaars, "Subspace Alignment Based Domain Adaptation for RCNN Detector," in *Proc. BMVC*, 2015.

[42] Y. Chen, W. Li, C. Sakaridis, D. Dai, and L. Van Gool, "Domain Adaptive Faster R-CNN for Object Detection in the Wild," in *Proc. CVPR*, 2018, pp. 3339–3348.

[43] J. Hoffman *et al.*, "CyCADA: Cycle-Consistent Adversarial Domain adaptation," in *Proc. ICML*, 2018, pp. 3162–3174.

[44] A. Dundar, M.-Y. Liu, T.-C. Wang, J. Zedlewski, and J. Kautz, "Domain Stylization: A Strong, Simple Baseline for Synthetic to Real Image Domain Adaptation," 2018. [Online]. Available: arXiv:1807.09384.

[45] J. Y. Zhu, T. Park, P. Isola, and A. A. Efros, "Unpaired Image-to-Image Translation Using Cycle-Consistent Adversarial Networks," in *Proc. ICCV*, 2017, pp. 2242–2251.

[46] X. Huang, M. Y. Liu, S. Belongie, and J. Kautz, "Multimodal Unsupervised Image-to-Image Translation," in *Proc. ECCV*, 2018, pp. 179–196.

[47] S. Ben-David, J. Blitzer, K. Crammer, A. Kulesza, F. Pereira, and J. W. Vaughan, "A theory of learning from different domains," *Mach. Learn.*, vol. 79, pp. 151–175, 2010.

[48] A. Singh, J. Sha, K. S. Narayan, T. Achim, and P. Abbeel, "BigBIRD : A Large-Scale 3D Database of Object Instances," in *Proc. ICRA*, 2014.

[49] K. Lai, L. Bo, X. Ren, and D. Fox, "A large-scale hierarchical multi-view RGB-D object dataset," in *Proc. ICRA*, 2011, pp. 1817–1824.

[50] C. Rother, V. Kolmogorov, and A. Blake, "'GrabCut' - Interactive Foreground Extraction using Iterated Graph Cuts," in *SIGGRAPH*, 2004.

[51] H. Noh, S. Hong, and B. Han, "Learning Deconvolution Network for Semantic Segmentation," in *Proc. ICCV*, 2015, pp. 1520–1528.

[52] J. Long, E. Shelhamer, and T. Darrell, "Fully Convolutional Networks for Semantic Segmentation," in *Proc. CVPR*, 2015, pp. 3431–3440.

[53] P. Ammirato, P. Poirson, E. Park, J. Kosecka, and A. C. Berg, "A Dataset for Developing and Benchmarking Active Vision," in *Proc. ICRA*, 2017, pp. 1378–1385.

[54] G. Georgakis, M. A. Reza, A. Mousavian, P. H. Le, and J. Kosecka, "Multiview RGB-D dataset for object instance detection," in *Proc. 3DV*, 2016, pp. 426–434.

[55] T. Y. Lin *et al.*, "Microsoft COCO: Common Objects in Context," in *Proc. ECCV*, 2014, pp. 740–755.

[56] Y. Cho, J. Jeong, Y. Shin, and A. Kim, "DejavuGAN: Multi-temporal Image Translation toward Long-term Robot Autonomy," in *Proc. ICRA Workshop*, 2018.

[57] S. Li, X. Xu, L. Nie, and T. S. Chua, "Laplacian-steered Neural Style Transfer," in *ACM MM*, 2017, pp. 1716–1724.

[58] M. Everingham, S. M. A. Eslami, L. Van Gool, C. K. I. Williams, J. Winn, and A. Zisserman, "The Pascal Visual Object Classes Challenge: A Retrospective," *Int. J. Comput. Vis.*, vol. 111, no. 1, pp. 98–136, 2014.

[59] N. Silberman, D. Hoiem, P. Kohli, and R. Fergus, "Indoor Segmentation and Support Inference from RGBD Images," in *Proc. ECCV*, 2012, pp. 746–760.

[60] L. van der Maaten and G. Hinton, "Visualizing Data using t-SNE," *J. Mach. Learn. Res.*, vol. 9, pp. 2579–2605, 2008.

[61] W.-G. Chang, T. You, S. Seo, S. Kwak, and B. Han, "Domain-specific Batch Normalization for Unsupervised Domain Adaptation," in *Proc. CVPR*, 2019.

[62] S. Cui, S. Wang, J. Zhuo, C. Su, Q. Huang, and Q. Tian, "Gradually Vanishing Bridge for Adversarial Domain Adaptation," in *Proc. CVPR*, 2020.

[63] K. Saito, Y. Ushiku, T. Harada, K. Saenko, "Strong-Weak Distribution Alignment for Adaptive Object Detection," in *Proc. CVPR*, 2019.

[64] D. Zhang, J. Li, L. Xiong, L. Lin, M. Ye, and S. Yang, "Cycle-Consisten Domain Adaptive Faster RCNN," *IEEE Access*, vol. 7, pp. 123903-123911, 2019.

[65] H.-K. Hsu et al., "Progressive Domain Adaptation for Object Detection," in *Proc. WACV*, 2020.

[66] N. Inoue et al., "Cross-Domain Weakly-Supervised Object Detection through Progressive Domain Adaptation," in *Proc. CVPR*, 2018.

[67] A. RoyChowdhury et al., "Automatic Adaptation of Object Detectors to New Domains using Self-training," in *Proc. CVPR*, 2019.

[68] F. Yu et al., "Unsupervised Domain Adaptation for Object Detection via Cross-Domain Semi-Supervised Learning," in *Proc. CVPR Workshop*, 2020.

[69] L. Song et al., "Unsupervised Domain Adaptive Re-identification: Theory and practice," *Pattern Recognit.*, 2020.

[70] I. Shin et al., "Two-phase Pseudo Label Densification for Self-training based Domain Adaptation," in *Proc. ECCV*, 2020.

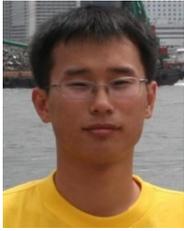

**WOO-HAN YUN** received his B.S. degree in electronic and electrical engineering from SungKyunKwan University, Suwon, South Korea, in 2004 and the M.S. degree in computer science and engineering from POSTECH, Pohang, South Korea, in 2006. He is currently a Ph.D. student at KAIST, Daejeon, Korea.

He joined ETRI (Electronics and Telecommunications Research Institute) in 2006. His current research interests include object detection, recognition, and human–robot interaction.

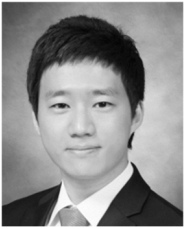

**TAEWOO KIM** received his B.S. degree from the School of Mechatronics Engineering, Chungnam National University, Daejeon, South Korea, in 2011, where he is currently pursuing his Ph.D. degree with the Department of Computer Software and Engineering, Korea University and Science and Technology, Daejeon, South Korea.

His research interests include robot kinematic control, motion retargeting, and reinforcement learning in robotic systems.

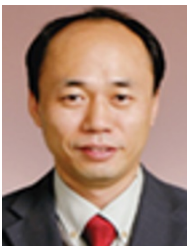

**JAEYEON LEE** received his Ph.D. degree from Tokai University (Japan) in 1996.

He has been a research scientist at ETRI (Electronics and Telecommunications Research Institute) since 1986. His research interests include robotics, pattern recognition, and computer vision.

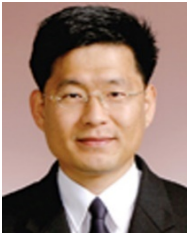

**JAEHONG KIM** received his Ph.D. degree from Kyungpook National University, Daegu, South Korea, in 2006.

He has been a research scientist at ETRI, Daejeon, Rep. of Korea since 2001. His research interests include socially assistive robotics for elder care, human–robot interaction, and gesture/activity recognition.

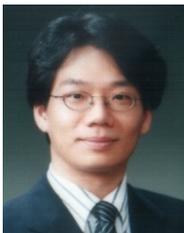

**JUNMO KIM** (S'01–M'05) received his B.S. degree from Seoul National University, South Korea, in 1998, and M.S. and Ph.D. degrees from Massachusetts Institute of Technology, in 2000 and 2005, respectively.

From 2005 to 2009, he was with the Samsung Advanced Institute of Technology, South Korea. He is currently a tenured associate professor with the Department of Electrical Engineering, KAIST, South Korea. His current research interests include image processing, computer vision, statistical signal processing, machine learning, and information theory.